%% file: top.tex
\pdfoutput=1
\documentclass[runningheads]{llncs}

\usepackage{epsfig}
\usepackage{graphicx}
\usepackage{amsmath}
\usepackage{amssymb}
\usepackage{color}
\usepackage{multirow}
\usepackage{eso-pic}
\usepackage[table,xcdraw]{xcolor}
\usepackage{comment}
\usepackage{cite}

\usepackage{wrapfig}
\usepackage[T1]{fontenc}
\usepackage{caption}
\usepackage{subcaption}

\usepackage{xspace}

\graphicspath{{../}}

\makeatletter
\DeclareRobustCommand\onedot{\futurelet\@let@token\@onedot}
\def\@onedot{\ifx\@let@token.\else.\null\fi\xspace}

\def\eg{\emph{e.g}\onedot} 
\def\ie{\emph{i.e}\onedot}

\makeatother

\usepackage[pagebackref=true,breaklinks=true,letterpaper=true,colorlinks,bookmarks=false]{hyperref}

\newcommand*\samethanks[1][\value{footnote}]{\footnotemark[#1]}

\usepackage{parskip}
\begin{document}
\pagestyle{headings}
\mainmatter
\def\ECCVSubNumber{2888}  

\title{Interactive Annotation of 3D Object Geometry using 2D Scribbles} 

\titlerunning{Interactive Annotation of 3D Object Geometry using 2D Scribbles}
%
\author{Tianchang Shen\inst{1,2}\thanks{equal contribution}\and
Jun Gao\inst{1,2,3}\samethanks \and
Amlan Kar\inst{1,2,3} \and
Sanja Fidler\inst{1,2,3}}
\authorrunning{T. Shen et al.}
%
\institute{$^1$University of Toronto, $^2$Vector Institute, $^3$Nvidia\\
\email{\{shenti11, jungao, amlan, fidler\}@cs.toronto.edu}}

\maketitle

\begin{abstract}
Inferring detailed 3D geometry of the scene is crucial for robotics applications, simulation, and 3D content creation. However, such information is hard to obtain, and thus very few datasets support it. In this paper, we propose an interactive framework for annotating 3D object geometry from both point cloud data and RGB imagery. The key idea behind our approach is to exploit strong priors that humans have about the 3D world in order to interactively annotate complete 3D shapes. Our framework targets naive users without artistic or graphics expertise. We introduce two simple-to-use interaction modules. First, we make an automatic guess of the 3D shape and allow the user to provide feedback about large errors by drawing scribbles in desired 2D views. Next, we aim to correct minor errors, in which users drag and drop mesh vertices, assisted by a neural interactive module implemented as a Graph Convolutional Network. Experimentally, we show that only a few user interactions are needed to produce good quality 3D shapes on popular benchmarks such as ShapeNet, Pix3D and ScanNet. We implement our framework as a web service and conduct a user study, where we show that user annotated data using our method effectively facilitates real-world learning tasks. Web service: \href{http://www.cs.toronto.edu/~shenti11/scribble3d}{http://www.cs.toronto.edu/$\sim$shenti11/scribble3d}.
\end{abstract}

\input{intro.tex}
\input{related_work.tex}

\input{method.tex}
\input{exp.tex}
\input{user_study.tex}

\input{conclude.tex}

\bibliographystyle{splncs04}

\input{top.bbl}
\end{document}

%% file: intro.tex
\section{Introduction}

3D scene understanding is a crucial component of numerous robotic applications such as autonomous driving, household robots, and delivery drones. 
In order to successfully plan its next move, an agent needs to infer 3D geometry of both the scene and relevant objects. Furthermore, the inferred 3D geometry should be sufficiently detailed to afford fine-grained interaction such as manipulation and grasping. 
We stress that reasoning about the visible scene alone is insufficient: objects occupy physical space in the world, extending beyond what is visible, which needs to be taken into account for downstream tasks. Reconstructing detailed and complete geometry of each object is also important for simulation, where the goal is to convert scanned point cloud scenes into interactive virtual environments to train artificial agents in, prior to real world deployment.  

\begin{figure}[t!]
    \begin{center}
        \includegraphics[width=0.85\linewidth]{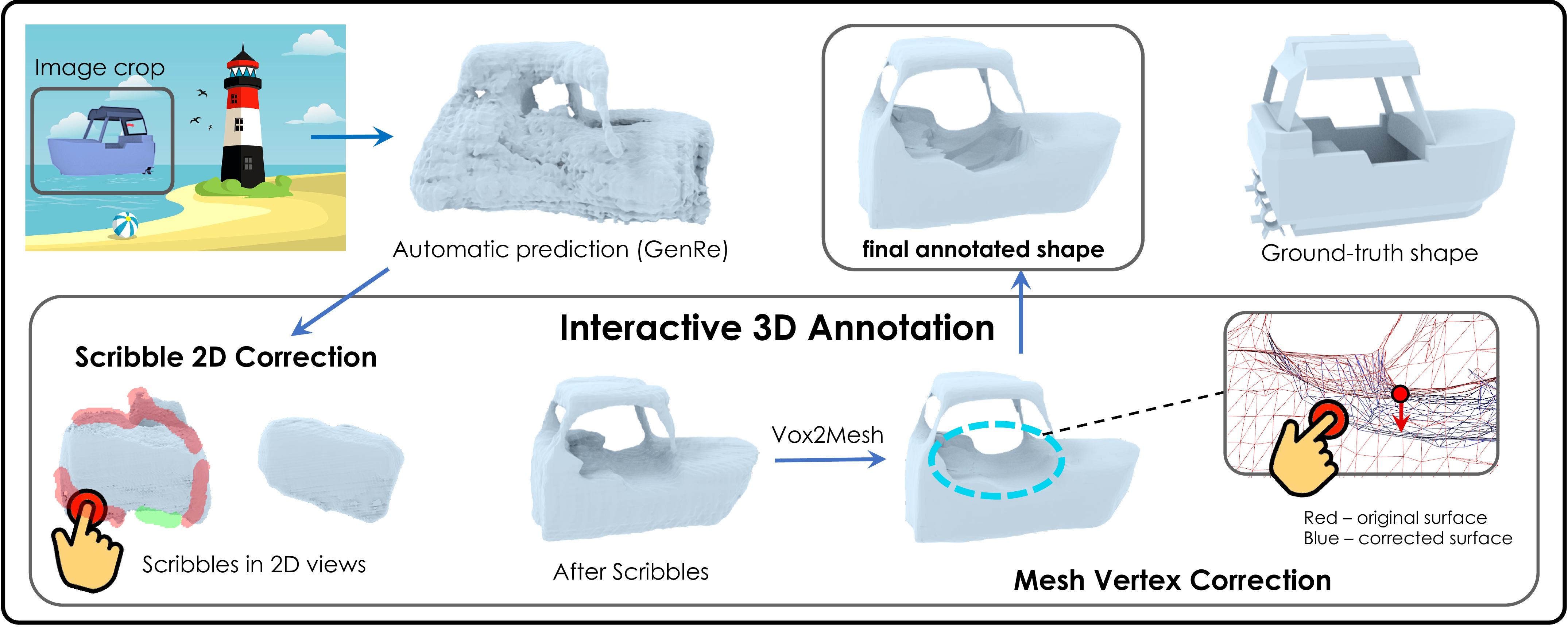}
    \end{center}
    \caption{Interactive 3D Annotation Framework: Our framework has two interaction modules: {\bf 1)} {\bf large errors}: user provides feedback by drawing scribbles in desired 2D views -- we exploit this to predict a better shape, {\bf 2)} {\bf small errors}:  user drags \& drops mesh vertices, and we repredict the mesh locally. Invisible object regions are up to user's priors/imagination.}
    \label{fig:teaser}
\end{figure}
We need datasets in order to infer complete 3D object geometry from partial observations (point clouds or imagery), 
 which have been collected with two dominant approaches. In~\cite{ScanDataset16}, the authors exhaustively scan each individual object, followed by a point-registration method. However, detailed scanning is time consuming and limited to static scenes and objects that can be scanned from all possible views -- which may require lifting the object from its support surface. 
The second approach employs fitting CAD models to segmented point clouds~\cite{dai2017scannet,NYUCAD,Apollo18} or RGB images~\cite{xiang2016objectnet3d,IKEA13}.  This is an undesirable alternative as it requires a large and diverse collection of CAD models to faithfully match input shapes. Such collections are not always available for any object class.

In this paper, we propose an interactive framework for annotating 3D object geometry from both point clouds and RGB images (Fig.~\ref{fig:teaser}). At its core, we exploit the extensive prior knowledge in humans to complete (annotate) objects. Since 3D CAD model creation typically requires artistic expertise, our key contribution lies in an interactive learning-based approach that enables a graphics layman to produce good quality 3D meshes that fit (partial) 3D observations.

Our annotation framework is composed of two stages. First, we employ existing methods to produce an initial guess of the 3D shape. To leverage the fact that 2D interactions are easier on existing digital media than 3D, we introduce a simple interface to correct large 3D errors by scribbling corrections in one or several desired 2D views. We propose a neural architecture that uses these scribbles to produce an improved prediction of the 3D shape. 
Next, for final fine-grained corrections, the annotator makes edits directly to 3D vertices of the object mesh using a drag-and-drop interface. Our 3D Graph Convolutional Network takes this correction into account and aims to re-predict refined local geometry, while ensuring that the majority of the object geometry remains intact.

We demonstrate that our interactive approach, \textbf{1)} is able to generate high-quality meshes from single images on the ShapeNet~\cite{shapenet} and Pix3D~\cite{pix3d} datasets, even on unseen object categories with very few interactions from an user. The annotated shapes can facilitates real-world learning tasks, 
\textbf{2)} comparing favourably to existing annotation strategies such as those employing CAD model retrieval and
\textbf{3)} can be used for full-scene annotation on the popular ScanNet dataset~\cite{ScanDataset16}.

%% file: related_work.tex
\section{Related Work}

\textbf{3D Reconstruction:}
The problem of recovering 3D shape from RGB-D or RGB images is challenging due to the inherent ambiguity in going from a lower to a higher dimensional space. Existing learning-based works can be categorized into four different categories based on output representation~\cite{kaolin2019arxiv}: voxel~\cite{choy20163d,multiview}, point cloud~\cite{fan2017point}, mesh~\cite{pixel2mesh,meshrcnn,dib-r} and signed distance field~\cite{occnet,disn}. In practice, by design, these approaches 
memorize the classes seen during training and exhibit limited generalization ability to unseen classes. Recently, GenRe~\cite{genre} proposed to exploit spherical maps in order to perform depth completion which has been shown to achieve better generalization to unseen classes. Our framework employs GenRe for its generalization ability, but is not limited to it, \ie other methods can be used. Our key contribution is in making these methods interactive.

\textbf{Interactive 3D Modeling} is a long studied topic in computer graphics~\cite{3-sweep,shtof2013geosemantic,gingold2009structured,delanoy20183d}. Image-based methods assist the user to snap parametric primitives (\eg generalized cylinders, ellipsoids) to image contours~\cite{3-sweep,shtof2013geosemantic,gao2019deepspline}. A combination of such primitives forms the final shape.~\cite{gingold2009structured} further allows users to indicate higher-level semantic information, \eg equal length/angles, symmetries. The major drawback of these methods is that they require the user to mentally imagine how to decompose 3D shape into primitives and perform their composition. This is not an easy task for a naive user. 
In our paper, we propose a learning-based interactive approach that assists a graphics layman to annotate complex shapes. 

\textbf{Data-driven approaches} have also been proposed to create 3D models. Early works mainly perform retrieval from a database, by either retrieving a whole 3D shape~\cite{funkhouser2003search}, or procedurally obtain predefined assemblies~\cite{lee2008sketch,xie2013sketch}.~\cite{lipson2000conceptual}  proposed to learn the correlation between 2D sketches and 3D shape. Recently,~\cite{delanoy20183d} utilized a deep network to learn this correlation, and users are able to modify the 3D shape via 2D sketches. However, drawing detailed sketches  requires some level of artistic expertise.~\cite{liu2017interactive} extend the idea of ~\cite{zhu2016generative} into 3D by utilizing a GAN to learn a latent representation of 3D shapes. User's edits are mapped to this latent space and the network regenerates a shape. This approach loses control over the generation and the network easily neglects user's corrections.

%% file: method.tex
\section{Interactive 3D Annotation}

We now describe our framework for annotating 3D object. Key desired properties include: \textbf{1)} naturally incorporate user interactions, \textbf{2)} generalize to unseen object classes and \textbf{3)} produce high-quality shapes. We emphasize that generalization beyond training shapes is crucial to scale to the diversity in the real-world. Since our goal is to produce annotated data to power applications required to predict high quality 3D shapes (\eg VR/AR, simulation, manipulation in robotics), we want our framework to support annotation of detailed geometry. 
Furthermore, we want our interactive tool to be seamlessly usable by anyone, without assuming expertise in 3D modeling, which is key in undertaking large-scale annotation endeavours in the future. 

Our method supports annotating 3D geometry from RGB images and point clouds. It is designed to reconstruct a complete 3D object shape in a coarse-to-fine manner, incorporating annotator corrections at all stages. In particular, it consists of two modules. The \emph{Scribble Interaction Module} (SIM), allows the user to make dramatic changes to the annotated shape by drawing coarse scribbles. The user inspects the 3D shape by rotating it to any viewpoint, and makes corrections to the projected silhouette by drawing scribbles in 2D, which is used to re-predict the shape. The user iterates until the desired quality is achieved. 
Next, the \emph{Point Interaction Module} (PIM) allows the user to make fine corrections directly in 3D by moving vertices of the object's mesh to obtain fine-grained geometric details. We use this feedback to re-predict the local mesh geometry in order to minimize human effort in editing the final mesh. 


In Section~\ref{sec:vol-recons}, we describe our annotation setup, including the initial automatic prediction of the 3D shape. In Section~\ref{sec:scribble}, we introduce our Scribble Interaction Module. Finally, we describe our Point Interaction Module in Section~\ref{sec:point}. 

\subsection{Annotation Setup}
\label{sec:vol-recons}

We envision a human user annotating 3D shapes for possibly multiple objects in an image (or a point cloud scene). The user is expected to indicate which object to annotate by drawing a 2D mask overlaying the object. Note that this can be done very quickly by using existing interactive techniques~\cite{curvegcn,zian19levelset}. Given the selected object, we aim to predict and assist in the interactive annotation of an accurate 3D shape. Our framework can incorporate any existing 3D prediction network, and our main contribution is in making annotation interactive.

\subsubsection{Automatic 3D Shape Prediction:} We choose GenRe~\cite{genre} as our automatic prediction network, for its ability to generalize to beyond training shapes. We briefly summarize it here and refer readers to the original paper for details. 
GenRe uses three steps: 1) predicting a depth map from an RGB image (cropped to contain a single object), 2) converting the depth map into a spherical map and inpainting the missing depth information with a 2D CNN and 3) refining the 3D shape. The first module is optional and is used when the input is a monocular image. With access to a depth map or a point cloud, this step can be omitted. In the third module, the completed spherical depth map from the second module is back-projected into a 3D occupancy grid, which is further refined to a volume of size $128^3$ using a 3D-UResNet (visualization is provided in the Appendix).

We note that the resulting shape is viewer-centric, allowing us to easily overlay it onto sensor inputs (\eg RGB image, point cloud), which is useful for the user tasked to provide corrections to the predicted shape. When depth is available, we can place the object in camera coordinates, \ie into the world scene. 

\subsection{Scribble Interaction Module}
\label{sec:scribble}
The Scribble Interaction Module (SIM) helps correct major errors in the initial 3D shape prediction. It is purposefully designed to be 2D view-based, targeting an average 3D illiterate user. Unless specially trained to craft 3D content such as artists or graphics experts, a naive user is known to be better at 2D editing. SIM then learns to propagate 2D corrections to 3D in order to refine the shape.

\subsubsection{Annotation Setup:}
\label{sec:sim_anno_setup}
\begin{figure*}[t]
\begin{minipage}{0.13\linewidth}
        \centering
        \includegraphics[width=0.83\linewidth]{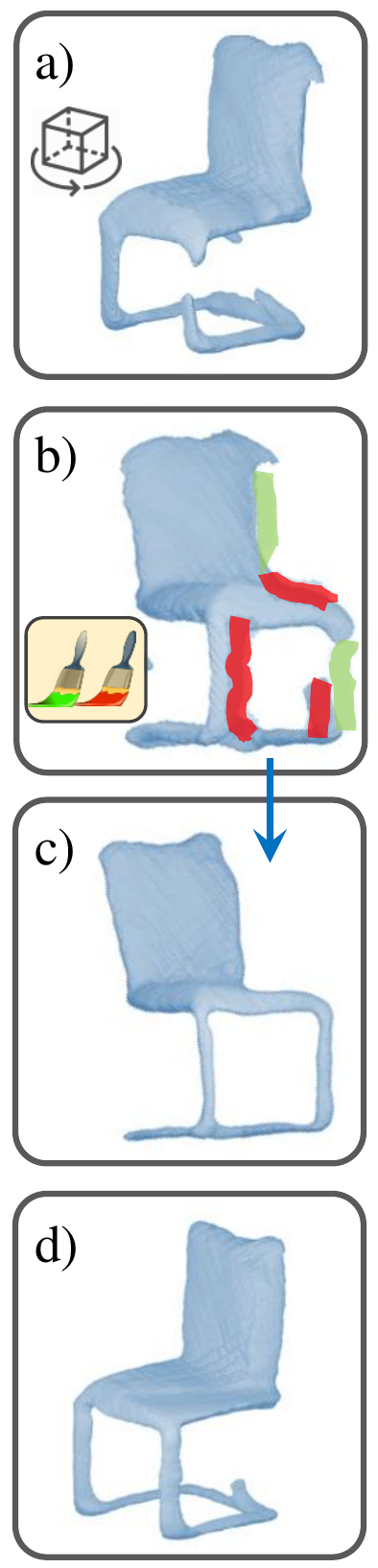}
   \caption{2D Scribble Annotation.}
           \label{fig:simtool}
\end{minipage}
\hspace{2.8mm}
\begin{minipage}{0.84\linewidth}
    \begin{center}
        \includegraphics[width=0.98\linewidth]{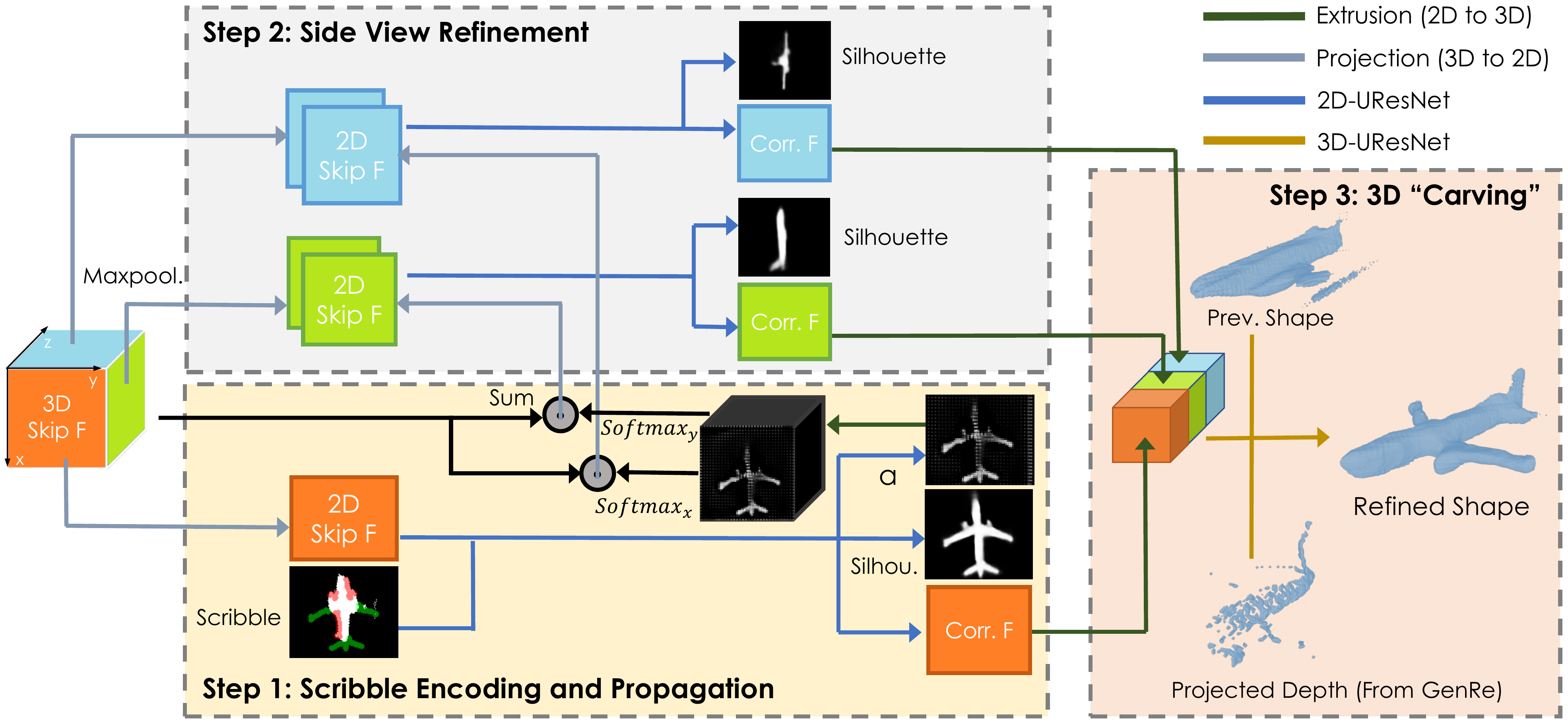}
    \end{center}
   \caption{{\bf Our Scribble Interaction Module}: We first get 3D skip features from GenRe (left cube), which we rotate and project to the user's view and its orthogonal views. User draws scribbles in the 2D view to indicate errors in the predicted shape. Our network encodes and propagates scribbles to orthogonal views. 
   Lastly, we concatenate refined features from all three orthogonal views, along with the shape from the previous step and the projected depth, to predict a new refined shape.}
\label{fig:simmodel}
\end{minipage}
\end{figure*}

To make edits to a shape (automatic prediction initially), the user views it in 3D and rotates it to any desired viewpoint. If a view with a major flaw is discovered, the user can indicate errors by drawing scribbles onto the (2D) projection (Figure~\ref{fig:simtool} provides an illustration). Following existing work on scribble-based image annotation~\cite{video_annotation}, we support ``additive'' scribbles to indicate false negative areas and ``deletion'' scribbles to indicate false positive areas. 
A scribble contains pixels obtained by dilating the trace of the mouse cursor with a kernel of a user-specified bandwidth, called the scribble width. When the user wants to add (or remove) fine details to the 3D shape, a small scribble width can be used. Nonetheless, this interaction is coarse in the sense that a scribble typically does not accurately trace high frequency regions.

\subsubsection{Scribble-based Neural Module:}
\label{sec:sim_arch}

Since the scribble is marked in 2D, there is an inherent ambiguity in 3D associated with 2D correction. We design a neural architecture that makes a best guess of which voxels in the 3D volume should be corrected based on 2D scribbles. The annotation module allows the user to iteratively rotate the refined shapes and indicate errors, until a satisfactory quality is achieved. Our architecture (Figure~\ref{fig:simmodel}) is based on the intuition that predicting 2D projections of a shape is likely an easier task compared to predicting a 3D occupancy grid. The network needs to infer how the refined silhouette of the shape would look in the view in which the scribbles were drawn, and other related views (\eg two orthogonal views), while exploiting depth information and user scribbles. The shape can then be obtained by 3D carving using the predicted silhouettes, and refining the ``carved'' shape with a simple architecture. 

\subsubsection{Overview of Architecture:} The model takes as input a 3D feature map (from GenRe) and rotates it to align with the user-specified view. We perform max-pooling along the $z$ axis to obtain a \emph{2D-view feature representation} of the input projected into user's view. Scribbles are encoded along with the projected feature map to obtain a new representation of the corrected viewpoint using a \emph{2D-scribble encoder}. In order to propagate this information to the orthogonal views, we similarly project GenRe's feature map onto the two other views, and concatenate information propagated from the user-specified view to each of the projected feature maps, using the \emph{scribble propagation module}. We decode these concatenated maps into the final view representations using a 2D-UResNet. To predict the final 3D shape, the \emph{3D carving module} performs an operation mimicking carving in the feature space to obtain a 3D volumetric representation which is decoded into a 3D occupancy grid. We provide details next. 

\subsubsection{2D-view Feature Representation:} 
Let $R$ denote the rotation that rotates the object into the user's viewpoint. We extract skip feature maps, $F\in \mathbb{R}^{C\times64\times64\times64}$, from the first and last 3D convolutional layers of GenRe's 3D-UResNet and rotate it to the user's 3D viewpoint via $R$. The feature map is upscaled by a factor of 2 using bilinear interpolation, denoted by $F_{v}\in \mathbb{R}^{C\times128\times128\times128}$. $F_{v}$ is passed through a single FC layer and is projected to three orthogonal views by max-pooling along $x$, $y$, and $z$ axes, \ie $F_{v}^{xy} = \text{maxpool}_z(F_v)$ (similar for $F_{v}^{yz}$ and $F_{v}^{xz}$). Pooling along $z$ mimics an orthographic projection into the user's view, while pooling along $x$ and $y$ projects into its orthogonal views. While more elaborate pooling operations could be performed mimicking perspective projection, we found the simpler approach to work well in practice. 

\subsubsection{Iterative Correction:}
Note that our interactive approach runs iteratively, allowing the user to draw scribbles in a different viewpoint each time. In the first correction round, we use 3D features from GenRe, while in all subsequent rounds we use features from SIM. However, since GenRe's and SIM's outputs differ, we choose to use 2 different weight matrices for the first vs other steps. The 3D refinement network weights are shared across all correction rounds.

\subsubsection{Scribble Encoder:} We encode positive/negative scribbles by concatenating them along with the initial (binary) silhouette into $S\in \mathbb{R}^{3\times128\times128}$. To process scribbles, we use a 2D-UResNet, that takes a concatenation of the projected 2D feature $F_{v}^{xy}$ and $S$ as input and predicts a refined silhouette $M^{xy}$ as well as refined features $F_{c}^{xy}$ in the user's view:
\begin{eqnarray}
F_s &=& \text{concat}\{F_{v}^{xy},\ S\},  \\
M^{xy}, F_{c}^{xy} &=&\text{2D-UResNet} (F_s)
\end{eqnarray}
$F_{c}^{xy}$ is used in the 3D carving module, while the mask prediction is used for an auxilliary loss function during training.

\subsubsection{Scribble Propagation Module}
aims to propagate scribble information from the user's view into orthogonal views. We utilize a 2D-UResNet to predict an attention map $\mathbf{A}^{xy} = \text{2D-UResNet}(F_s)\in\mathbb{R}^{128\times 128}$, which functions as a gate to silence or amplify features of the entire tube. 
We then extrude the attention map along $z$-axis: $\hat{\mathbf{A}}^{xy}\in\mathbb{R}^{128\times 128\times 128}$,
and perform softmax along different axes when propagating to the orthogonal views: 
\begin{eqnarray}
\hat{\mathbf{A}}^{yz}_{i,j,k} = \frac{\exp(\hat{\mathbf{A}}_{i,j,k}^{xy})}{\sum_{i}{\exp(\hat{\mathbf{A}}_{i,j,k}^{xy})}}, \hat{\mathbf{A}}^{xz}_{i,j,k} = \frac{\exp(\hat{\mathbf{A}}_{i,j,k}^{xy})}{\sum_{j}{\exp(\hat{\mathbf{A}}_{i,j,k}^{xy})}}
\end{eqnarray}
where $i,j,k$ denote 3D tensor indices. We compute the weighted feature map:
\begin{eqnarray}
\hat{F}_{v}^{yz}  = F_{v} \circ \hat{\mathbf{A}}^{yz}, \hat{F}_{v}^{xz}  = F_{v} \circ \hat{\mathbf{A}}^{xz},
\end{eqnarray}
where $\circ$ denotes the Hadamard product. $M^{yz}, F_{c}^{yz}, M^{xz}, F_c^{xz}$ are predicted in a similar manner as in the user's view, but taking projected features from both $F_{v}$ and the weighted feature map. For the yz-plane features (similar for xz),
\begin{eqnarray}
 M^{yz}, F_{c}^{yz} &=&\text{2D-UResNet} (\text{concat}\{\sum_x(\hat{F}_{v}^{yz}), F_{v}^{yz}\})
\end{eqnarray}

\subsubsection{3D Carving Module:}
 The refined view features are backprojected to 3D by extruding $F_{c}^{xy}$,  $F_{c}^{yz}$ and  $F_{c}^{xz}$ along their orthogonal axes. A simple refinement network takes the concatenation of the three extruded feature maps, previous occupancy grid and the projected depth map as input, and outputs a new volume as the final prediction. The projected depth map is included to preserve consistency with sensor observations. Note that we do not explicitly use the predicted silhouettes, but instead use the predicted feature maps. We argue that the feature maps preserve more information and thus are better for 3D refinement, while predicting and supervising silhouettes helps learn better features.


\subsection{Point Interaction Module}
\label{sec:point}

A voxelgrid with limited resolution lacks fine geometric details. We thus convert the predicted occupancy grid to a triangular mesh and allow the annotator to drag-and-drop vertices, one at a time, and employ the point interaction module to locally refine the mesh based on user feedback. To convert the occupancy grid to a mesh, a Vox2Mesh method similar to~\cite{meshrcnn, pixel2mesh} is used, with minor but important modifications that yield visibly better meshes (details in suppl.). \begin{wrapfigure}[7]{r}{0.4\linewidth}
\centering
\includegraphics[width=1\linewidth]{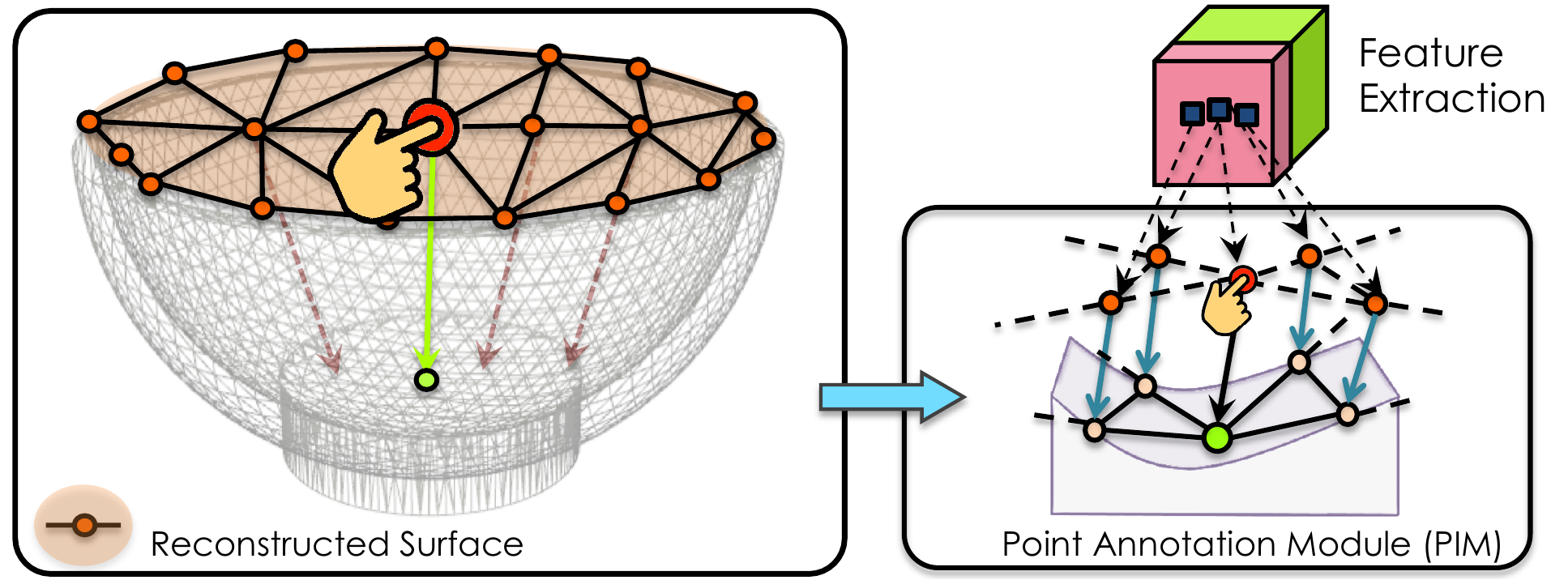}
\caption{PIM allows users to interactively edit a mesh.}
\label{fig:point_annot}
\end{wrapfigure}

\subsubsection{Point-based Neural Module} takes the user's vertex correction and employs a GCN to repredict its neighboring vertices. As shown in the Fig.~\ref{fig:point_annot}, we use a similar network architecture to~\cite{curvegcn}, which is designed for interactive 2D curve correction. In particular, it performs: 
\vspace{+2mm}
\begin{eqnarray}
f_i &=& \text{concat}(F(p_i); p_i; q_{i} - p_i),\\
\Delta p_j &=& \text{GCN} (f_1,f_2, \cdots, f_N; G),
\end{eqnarray}
where $F(p_i)$ represents the feature tilinearly sampled from 3D feature map with position $p_{i}$, which is corrected to $q_i$, $N$ is the number mesh vertices, and $G$ is the topology of the mesh. $\Delta p_j$ represents the predicted offset for each local point $p_j$. We restrict movement to only neighbours along paths of length $l$ from the edited vertex. It is trained using similar loss as MeshRCNN~\cite{meshrcnn} (see appendix).


%% file: exp.tex
\section{Experiments}

\label{sec:exp}

In this section, we provide extensive evaluation of our model. To demonstrate the model's capability as an annotation tool, we report interactive single image shape reconstruction results on ShapeNet for both seen and unseen classes in Sec.~\ref{sec:shapenet}. We further evaluate our full pipeline on the challenging Pix3D~\cite{pix3d}, ScanNet~\cite{dai2017scannet} and Scan2CAD~\cite{scan2cad} dataset to show the generalizability of our method to real-world sensor inputs in Sec.~\ref{sec:real-image}. 
In addition, we provide quantitative analysis comparing our method to CAD Retrieval using ShapeNet models to annotate Pix3D instances in Sec.~\ref{sec:real-image}. We show ablation studies comparing SIM with other network architectures in Sec.~\ref{sec:analysis}. Finally, we annotate real data in a user study and show results of models trained on this data in Sec.~\ref{sec:user}.


\begin{table*}[!t]
\resizebox{\textwidth}{!}{
\begin{tabular}{|l|l|c|c|c|c|c|c|c|c|c|c|c|}

\hline
\multicolumn{2}{|l|}{\multirow{2}{*}{}} & \multirow{2}{*}{Seen Classes} & \multicolumn{10}{c|}{Unseen Classes} \\ \cline{4-13} 
\multicolumn{2}{|l|}{}              &                                                                            & Bch   & Vsl & Rfl  & Sfa  & Tbl  & Phn  & Dsp   & Spk             & Lmp  & Avg           \\ \hline
\multirow{3}{*}{Chamfer Dist.}   & GenRe                 & 0.1093                                             &0.1158&0.1477&0.1570&0.1140&0.1572&0.1522& 0.1921& 0.1738          &0.2667&0.1640\\ \cline{2-13}
                                    & + Scribble Annot. & 0.0503                                             &0.0435&0.0517&\textbf{0.0289}&0.0465&0.0565&\textbf{0.0315}& \textbf{0.0461}& \textbf{0.0630} &0.1190&\textbf{0.0541}\\ \cline{2-13}
                                    & + Vox2Mesh            & \textbf{0.0474}                                    &\textbf{0.0412}&\textbf{0.0474}&0.0301&\textbf{0.0452}&\textbf{0.0545}&0.0345& 0.0499& 0.0681          &\textbf{0.1178}&0.0543\\ \hline
\multirow{3}{*}{F1 Score}           & GenRe                 & 0.8392                                             &0.8211&0.7270&0.6837&0.8368&0.7221&0.6993& 0.6014& 0.6832          &0.4927&0.6963\\ \cline{2-13} 
                                    & + Scribble Annot. & 0.9743                                             &0.9806&\textbf{0.9787}&0.9916&0.9629&0.9629&0.9871& \textbf{0.9774}& \textbf{0.9490} &0.8307&0.9578\\ \cline{2-13} 
                                    & + Vox2Mesh            & \textbf{0.9757}                                    &\textbf{0.9849}&0.9780&\textbf{0.9854}&\textbf{0.9858}&\textbf{0.9642}&\textbf{0.9885}& 0.9714& 0.9387          &\textbf{0.8320}&\textbf{0.9587}\\ \hline
\multirow{3}{*}{Normal Consist.} & GenRe                 & 0.6353                                             &0.6131&0.5640&0.4945&0.6266&0.6202&0.6029& 0.5632& 0.6030          &0.5004&0.5764\\ \cline{2-13} 
                                    & + Scribble Annot. & 0.7954                                             &0.7633&0.7522&\textbf{0.7454}&0.8429&0.8171&0.8907& 0.8506& 0.8371          &\textbf{0.7031}&0.8002\\ \cline{2-13} 
                                    & + Vox2Mesh            & \textbf{0.8128}                                    &\textbf{0.7905}&\textbf{0.7563}&0.7257&\textbf{0.8477}&\textbf{0.8360}&\textbf{0.8946}& \textbf{0.8692}& \textbf{0.8386} &0.7002&\textbf{0.8065}\\ \hline
\end{tabular}}
\vspace{+2mm}
\caption{Reconstruction results of the training classes and 9 unseen classes from ShapeNet. With our interactive method, we consistently improve and significantly improve over the automatic 3D reconstruction approach.}
\label{tab:shapenet}
\end{table*}

\begin{figure*}[ht!]
\centering
    \includegraphics[width=0.9\linewidth]{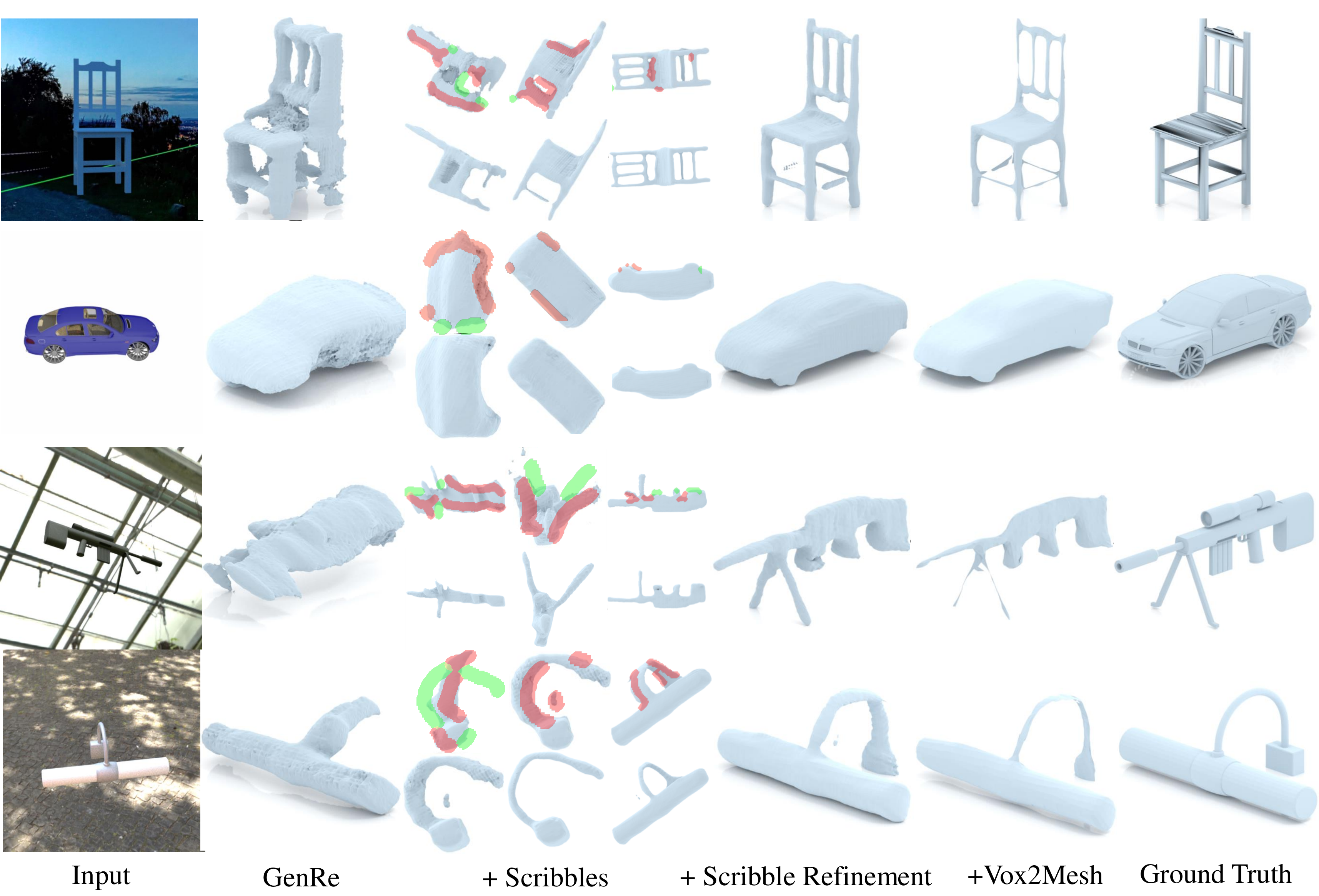}
\caption{Examples of ShapeNet objects annotated using our framework. First two rows are from seen classes while others are from unseen classes. Output shape from each intermediate step is shown from left to right, followed by GT mesh. Additive/deletion scribbles are shown in green/red.  Although the additive scribbles are drawn in 2D, SIM accurately infers the depth of added shape (e.g.\ bipods in the third example). With a few scribbles, our model predicts shape close to GT. Vox2Mesh further converts a discrete voxel grid to a fine mesh.}
\label{fig:examples}
\end{figure*}

\subsection{Experimental Settings}

For all experiments, we train our model on Car, Chair and Airplane, which are the three largest categories in ShapeNet~\cite{shapenet}. We use the same subset split and rendered images as in~\cite{genre}. 
For evaluation on real-world data, we use the Pix3D~\cite{pix3d} and ScanNet~\cite{dai2017scannet}. ScanNet is a richly annotated indoor dataset that provides RGB-D sequences, cameras, surface reconstructions, and instance-level semantic segmentation. It further provides CAD models aligned to the objects on Scan2CAD~\cite{scan2cad}. During inference, we use 5 correction steps for both interaction methods unless otherwise mentioned. More details are in appendix.\\

\begin{figure*}[ht!]
  \begin{minipage}[t]{0.32\linewidth}
  \captionsetup{width=.95\linewidth}
  \centering
      \includegraphics[height=3.1cm]{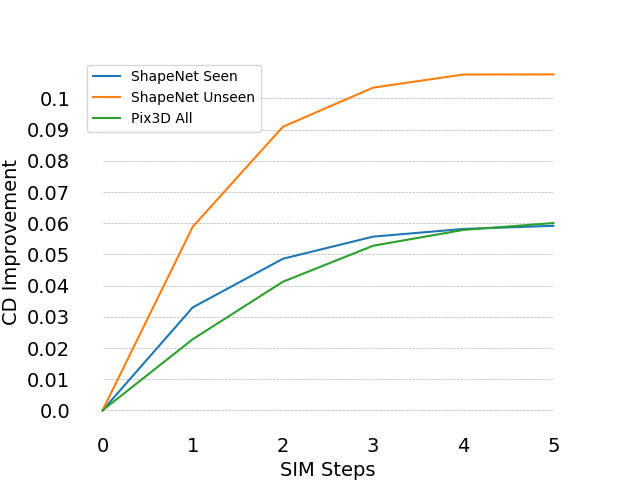}
  \caption{Improvements in Chamfer Dist vs num of scribble refin. steps. 
  }
  \label{fig:SIM_steps}
  \end{minipage}
  \begin{minipage}[t]{0.32\linewidth}
  \captionsetup{width=.95\linewidth}
  \centering
      \includegraphics[height=3.1cm]{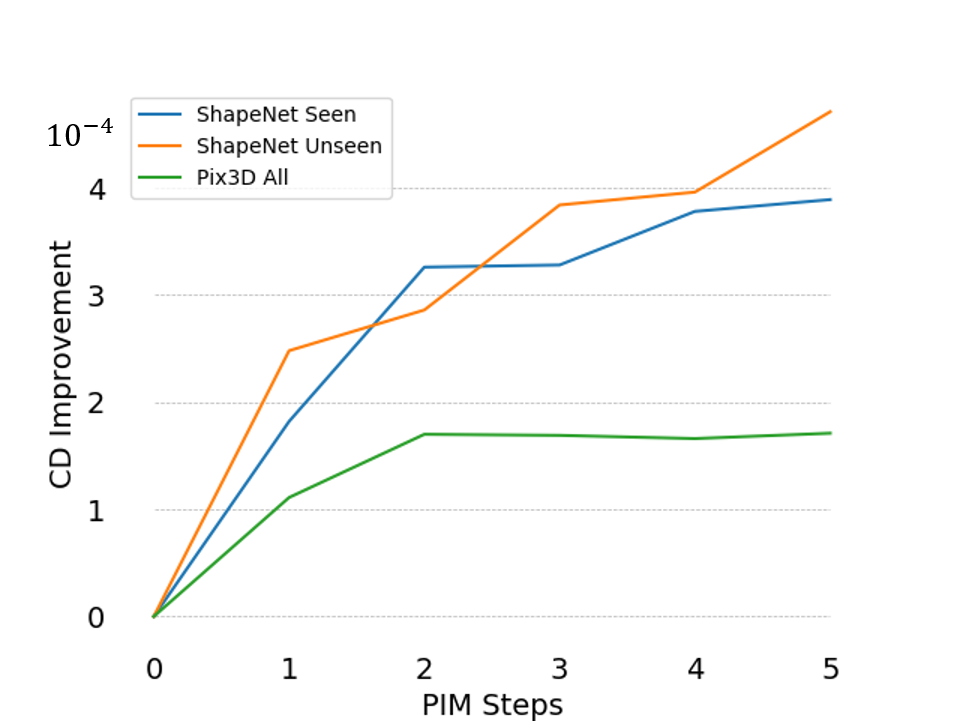}
  \caption{Improvements in Chamfer Dist vs. \# of point annotation steps.}
  \label{fig:PIM_steps}
  \end{minipage}
  \begin{minipage}[t]{0.32\linewidth}
  \captionsetup{width=.95\linewidth}
  \centering
      \includegraphics[height=3.1cm]{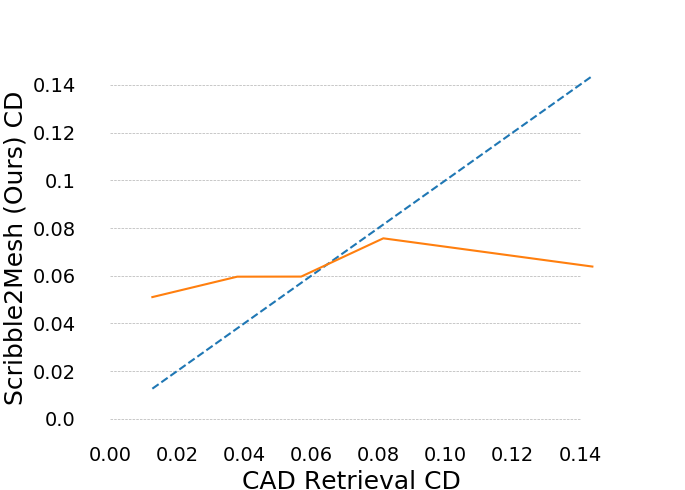}
  \caption{Chamfer Dist of the best-aligned CAD model vs. result obtained by our method. }
  \label{fig:cad_align}
  \end{minipage}
\end{figure*}

 \begin{figure}
    \hfill
    \begin{minipage}[c]{0.48\textwidth}
      \centering
      \includegraphics[width=\linewidth,trim=0 0 0 0,clip]{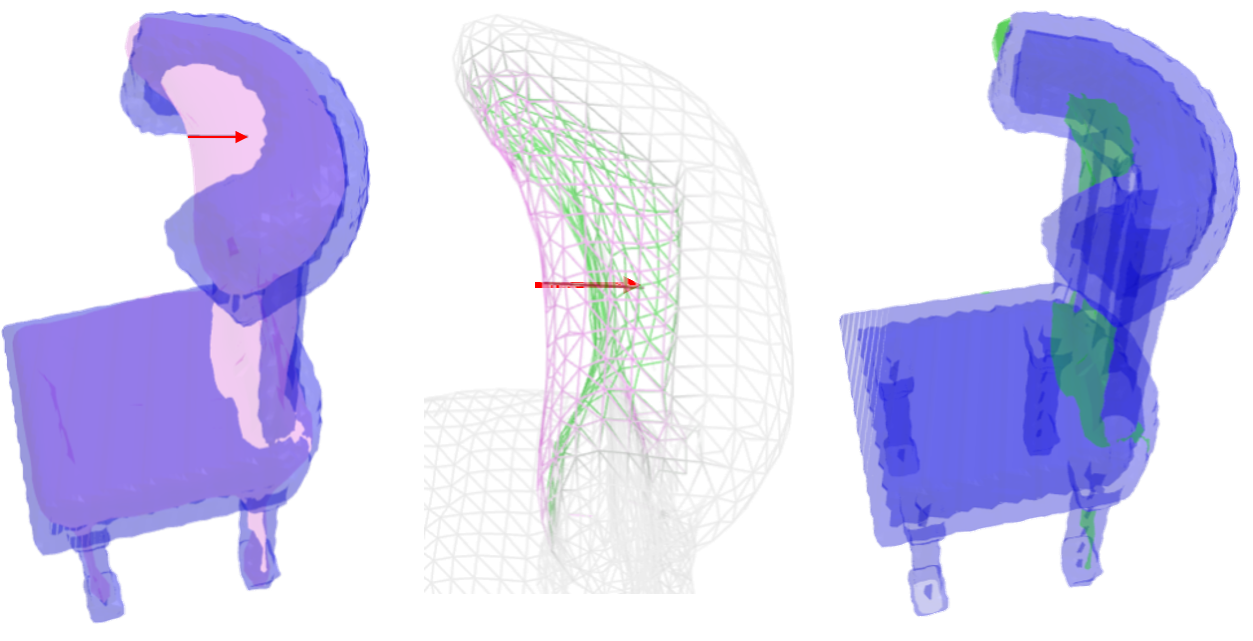}
    \end{minipage}\hfill
    \begin{minipage}[c]{0.47\textwidth}
      \caption{
  An example with PIM. The ground truth, initial prediction and refined prediction are colored in blue, pink and green, respectively. The red arrow represents correction for chosen vertex. The entire local surface is pushed towards the ground truth surface, demonstrating that PIM can annotate concave surfaces effectively.
      } \label{fig:pim_one_step}
    \end{minipage}
    \hfill
\end{figure}

\begin{table*}[t]
  \centering
  \resizebox{0.95\linewidth}{!}{%
  \begin{tabular}{|l|l|c|c|c|c|c|c|c|}
  \hline
                                      &                       & Chair           & Bed             & Bookcase & Desk & Sofa & Table & Wardrobe \\ \hline
  \multirow{3}{*}{Chamfer Distance}   & GenRe                 & 0.1161          & 0.1427          & 0.1167   &0.1526&0.0939&0.1735 &0.1128    \\ \cline{2-9} 
                                      & + Scribble Annotation & 0.0609          & \textbf{0.0696} & \textbf{0.0546}   &0.0748&0.0596&0.0968 &\textbf{0.0413}    \\ \cline{2-9} 
                                      & + Vox2Mesh            & \textbf{0.0588} & 0.0703          & 0.0575   &\textbf{0.0705}&\textbf{0.0579}&\textbf{0.0915} &0.0458    \\ \hline
  \multirow{3}{*}{F1 Score}           & GenRe                 & 0.8251          & 0.7483          & 0.8126   &0.7145&0.8957&0.6667 &0.8302    \\ \cline{2-9} 
                                      & + Scribble Annotation & 0.9549          & \textbf{0.9362} & \textbf{0.9660}   &0.9306&0.9720&0.8775 &\textbf{0.9851}    \\ \cline{2-9} 
                                      & + Vox2Mesh            & \textbf{0.9591} & 0.9358          & 0.9614   &\textbf{0.9345}&\textbf{0.9779}&\textbf{0.8808} &0.9819    \\ \hline
  \multirow{3}{*}{Normal Consistency} & GenRe                 & 0.6217          & 0.6479          & 0.5673   &0.6110&0.6541&0.6347 &0.6462    \\ \cline{2-9} 
                                      & + Scribble Annotation & 0.7358          & 0.7935          & 0.7022   &0.7807&\textbf{0.8160}&0.7784 &\textbf{0.8776}    \\ \cline{2-9} 
                                      & + Vox2Mesh            & \textbf{0.7587} & \textbf{0.8000} & \textbf{0.7037}   &\textbf{0.7990}&0.8147&\textbf{0.7969} &0.8628    \\ \hline
  \end{tabular}%
  }
  \vspace{+2mm}
  \caption{Reconstruction results on the  Pix3D Objects dataset.}
  \label{tab:pix3d}
\end{table*}

\subsection{ShapeNet Annotation}
\label{sec:shapenet}
\subsubsection{SIM:}
To evaluate SIM, we simulate user scribbles as in training (see appendix). We report results on both seen
and unseen classes in Tab ~\ref{tab:shapenet}. Improvement in Chamfer Distance as the corrections are sequentially applied is shown in Fig.~\ref{fig:SIM_steps}. With only coarse annotation, SIM significantly improves raw predictions under all metrics, and across all object categories with higher improvement in unseen classes. This demonstrates that our method generalizes to novel objects, making it suitable for the annotation task. We provide qualitative examples in Fig.~\ref{fig:examples}, demonstrating that with a few scribbles, our proposed module can significantly improve the initial prediction, bringing it very close to the ground-truth shape.

\subsubsection{PIM:} PIM is used to further refine the mesh (from Vox2Mesh). Following~\cite{curvegcn, polyrnnpp}, we use the same method to simulate the annotator to correct vertices in training. We report improvements in Chamfer Distance in Fig.~\ref{fig:PIM_steps}. As a local correction method, applying point annotation gradually improves the quality of 3D shape. Qualitative result in Fig.~\ref{fig:pim_one_step} shows that with 1 correction, PIM can effectively push the prediction to a concave surface, which cannot be annotated by scribbles.

\subsection{Annotating Real Scans}

\label{sec:real-image}
\subsubsection{Pix3D Annotation:}
We now evaluate our trained model on Pix3D~\cite{pix3d}. We do not fine-tune the model, and only run inference on this dataset. We test on un-occluded and un-truncated shapes and use ground-truth 3D shape to simulate scribbles. Quantitative results are shown in Tab.~\ref{tab:pix3d}. Our model achieves the same level of improvement across all Pix3D categories despite the domain gap between synthetic images rendered from ShapeNet models and real imagery. 

\begin{figure}[t]
\small
\begin{minipage}{0.14\textwidth}
\includegraphics[width=\linewidth, trim=300 400 300 100,clip]{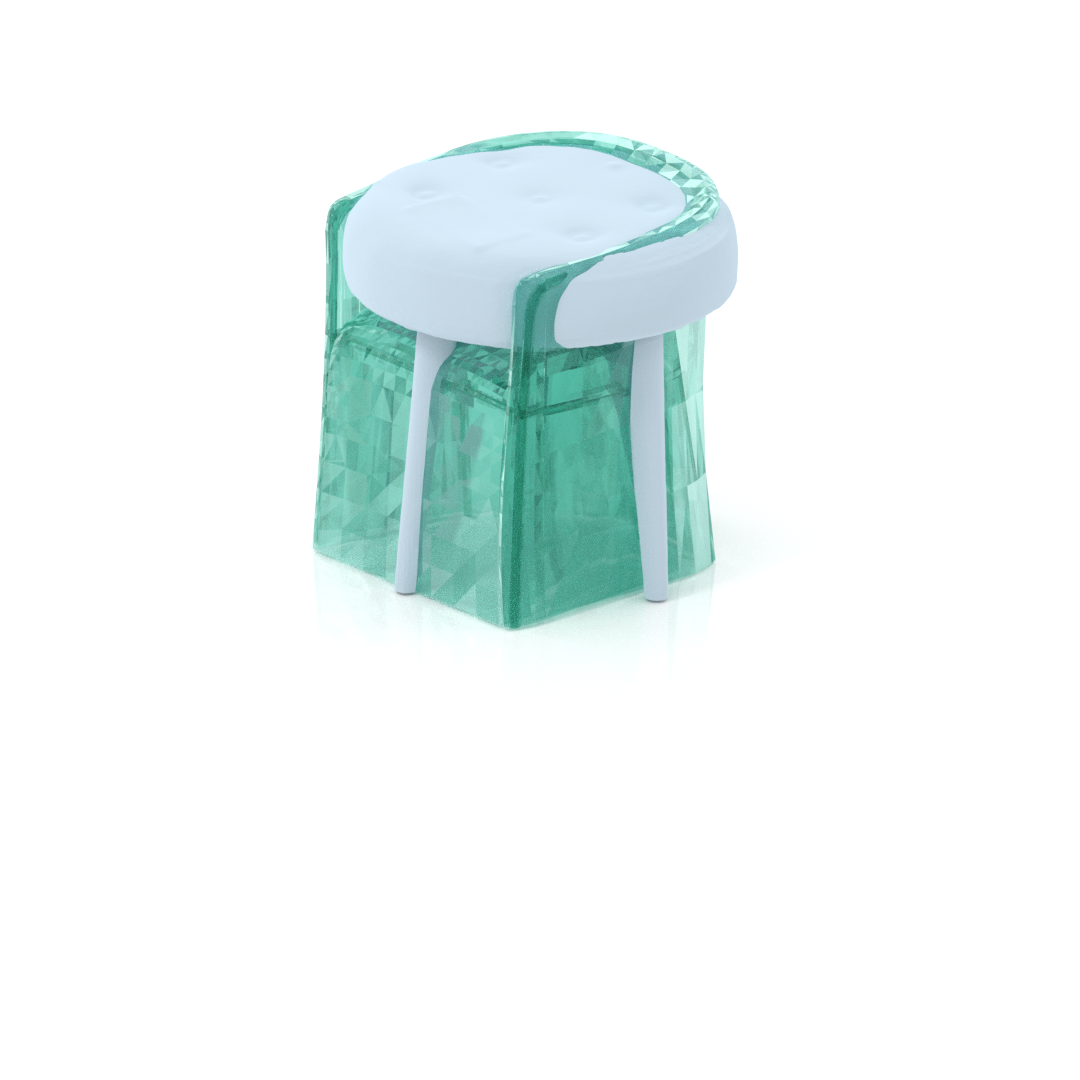}\\[-2mm]
{\centering
 \scriptsize (a)\par
}
\end{minipage}
\begin{minipage}{0.14\textwidth}
\includegraphics[width=\linewidth,trim=300 400 300 100,clip]{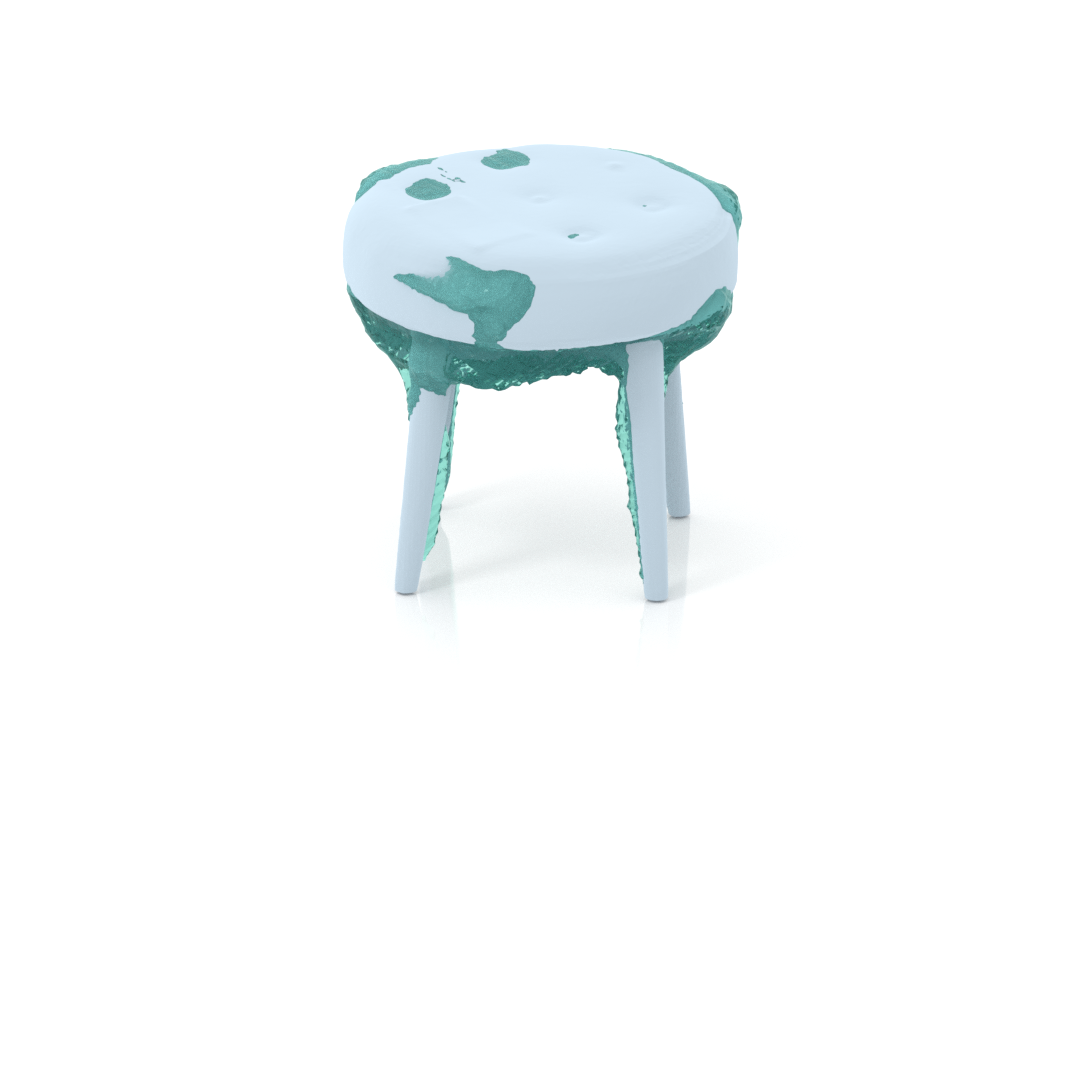}\\[-2mm]
{\centering
  \scriptsize (b)\par
}
\end{minipage}
\begin{minipage}{0.14\textwidth}
\includegraphics[width=\linewidth,trim=0 50 0 50,clip]{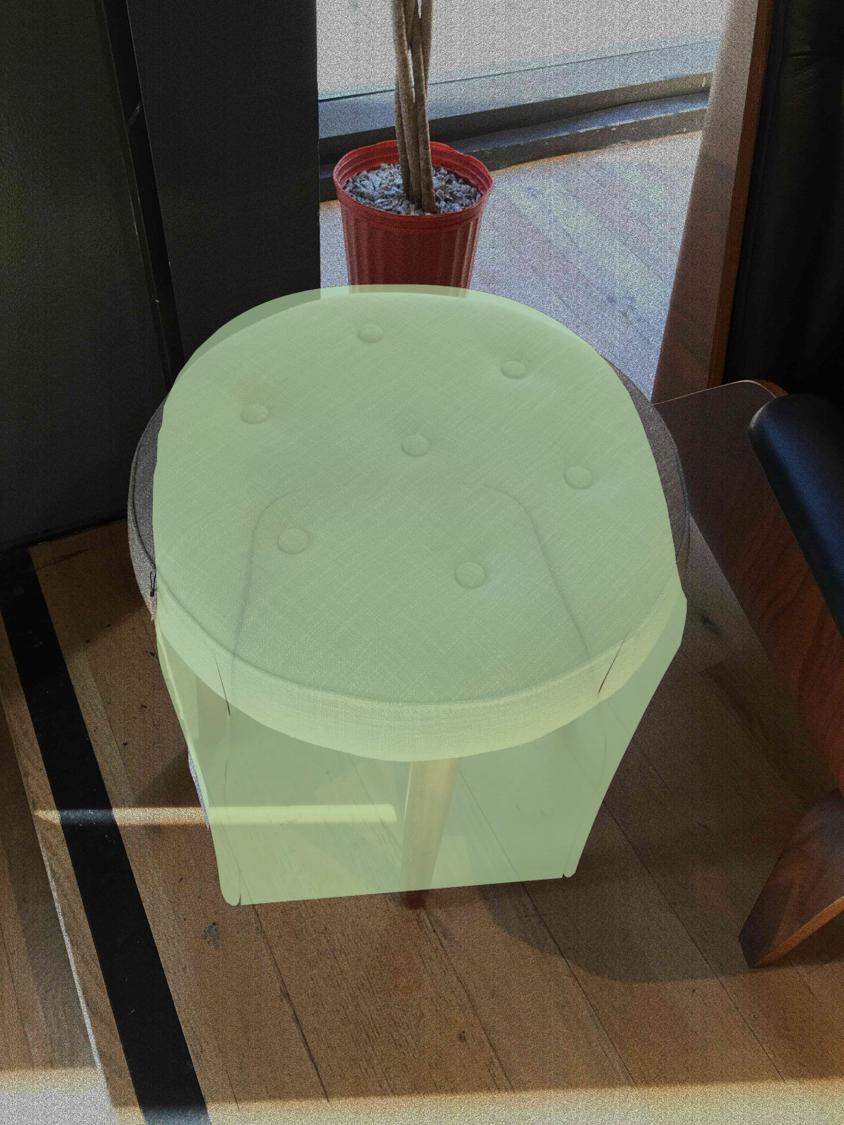}\\[-0mm]
{\centering
  \scriptsize (c)\par
}
\end{minipage}
\begin{minipage}{0.14\textwidth}
\includegraphics[width=\linewidth,trim=0 50 0 50,clip]{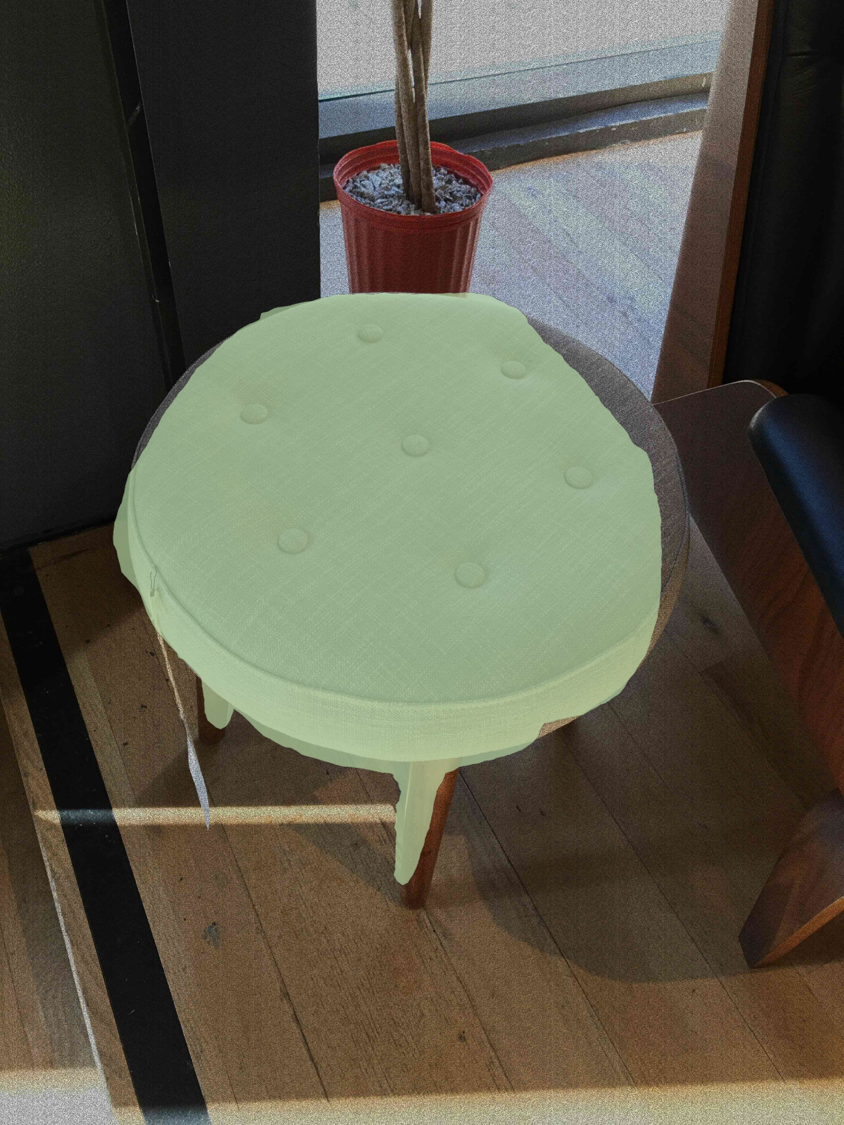}\\[-0mm]
{\centering
  \scriptsize (d)\par
}
\end{minipage}
\hspace{2mm}
  \begin{minipage}[c]{0.39\textwidth}
    \caption{
Example of CAD alignment (a) vs. our approach (b). Light blue shape is the ground truth. We render both shapes on the input image at c) and d). Our method is more consistent with both ground-truth shape and input.
    } \label{fig:1}
  \end{minipage}
\end{figure}

We compare with a common alternative approach for 3D annotation of retrieving and aligning CAD models from a large database~\cite{pix3d, scan2cad}. For fair comparison, we use our training and validation sets as the retrieval database. For retrieval, we pick the object with the minimum Chamfer Distance to the ground-truth CAD model (or scanned surface). Comparison on all 200 chair models in Pix3D is shown in Fig.~\ref{fig:cad_align}.
CAD retrieval is promising if the database contains similar objects, but it fails for unseen objects, which is crucial in practice. Our model overcomes this limitation and performs consistently regardless of the similarity to the training objects (Fig.~\ref{fig:1}). Our method achieves more faithful reconstruction to unseen objects, and can be used alongside CAD retrieval in a real application.

\subsection{Analysis}
\label{sec:analysis}

\subsubsection{Ablation Study:}
We explore two alternatives for scribble propagation. We could directly extrude 2D scribbles to 3D, or learn to encode the 2D scribbles in the correction view and extrude the learned feature directly to 3D (avoiding prediction of 2D side views in our model). The second approach is a generalization of the updater network in~\cite{delanoy20183d}, combined with volumetric aggregation in~\cite{iskakov2019learnable} to avoid repeated rotation of the feature space. We train these with the same training scheme and 3D U-ResNet architecture. The result is summarized in Table~\ref{tab:ablation}. Our proposed multi-view correction module outperforms both alternatives in all metrics, and the reconstruction is more plausible and complete (see Fig.~\ref{fig:ablation}).  



  \begin{minipage}{\textwidth}
    \hfill
  \begin{minipage}[b]{0.56\textwidth}
    \centering
        \includegraphics[width=0.8\linewidth, trim=80 40 80 0, clip]{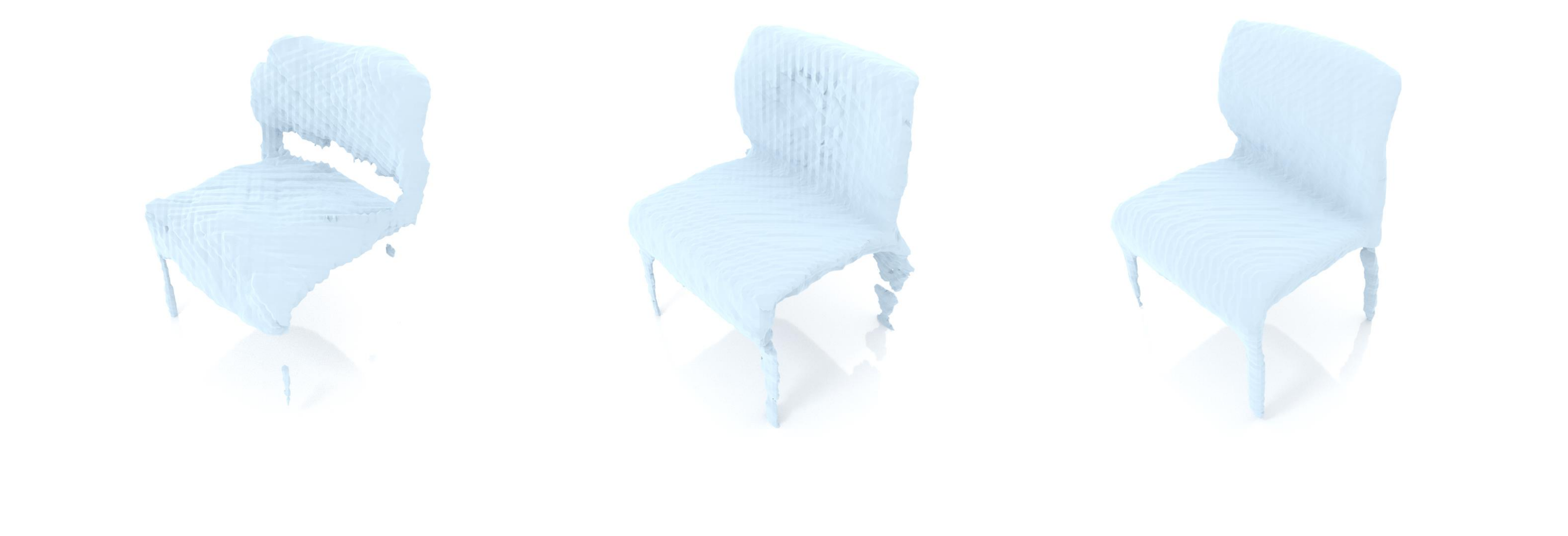}
    \captionof{figure}{(From left to right) The same chair annotated by 3D-scribble, single-view and multi-view SIM respectively.
    }
    \label{fig:ablation}
  \end{minipage}
  \hfill
  \begin{minipage}[b]{0.4\textwidth}
    \centering
    \resizebox{1\textwidth}{!}{
        \begin{tabular}{|l|c|c|c|c|c|c|}
        \hline
        \multirow{2}{*}{} & \multicolumn{3}{c|}{Seen} & \multicolumn{3}{c|}{Unseen} \\ \cline{2-7} 
                          & CD      & F1     & Normal & CD      & F1      & Normal  \\ \hline
        3D-Scribble       & 0.0628  & 0.9539 & 0.7729 & 0.0804  & 0.9109  & 0.7574  \\ \hline
        1-View            & 0.0555  & 0.9652 & 0.7795 & 0.0609  & 0.9499  & 0.7791  \\ \hline
        Multi-View        & \textbf{0.0535}  & \textbf{0.9694} & \textbf{0.7899} & \textbf{0.0589}  & \textbf{0.9538}         & \textbf{0.7838}        \\ \hline
        \end{tabular}%
    }
      \captionof{table}{Comparison of different SIM architectures.}
      \label{tab:ablation}

    \end{minipage}
    \hfill
  \end{minipage}

\subsubsection{Shape Creation with Scribbles:}
\begin{figure}
  \hfill
  \begin{minipage}[c]{0.55\textwidth}
    \includegraphics[width=\linewidth,trim=0 0 0 10,clip]{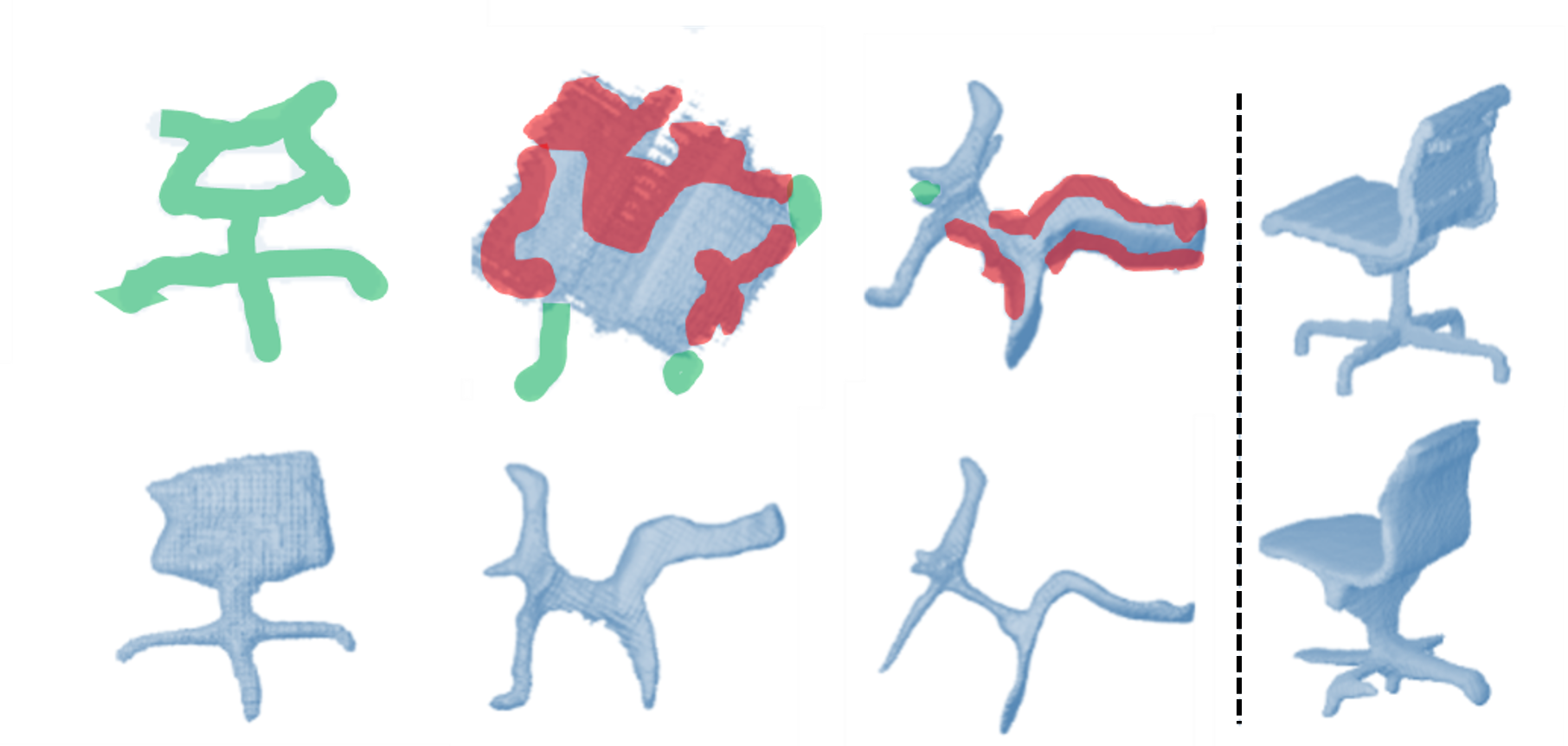}
  \end{minipage}\hfill
  \begin{minipage}[c]{0.42\textwidth}
    \caption{Shape creation from only scribble inputs. (top) Images on the left show scribbles, (bottom) object in correction view after refinement. The ground truth is shown on top right.
    } \label{fig:shape_creation}
  \end{minipage}
  \hfill
\end{figure}

Our model is capable of creating a plausible shape purely from scribble inputs, as shown in Fig.~\ref{fig:shape_creation}. This mimics the scenario where the user wants to create 3D shapes by simply drawing scribbles on a blank screen. As before, we allow the user to iteratively add scribbles from multiple views in order to obtain the desired shape. In the example,  the user is able to create a plausible full chair with only 3 drawing steps.

%% file: user_study.tex
\vspace{-2mm}
\subsection{User Study}

\label{sec:user}
\begin{wrapfigure}[9]{r}{0.3\linewidth}
  \centering
  \vspace{-2mm}
  \includegraphics[width=1\linewidth]{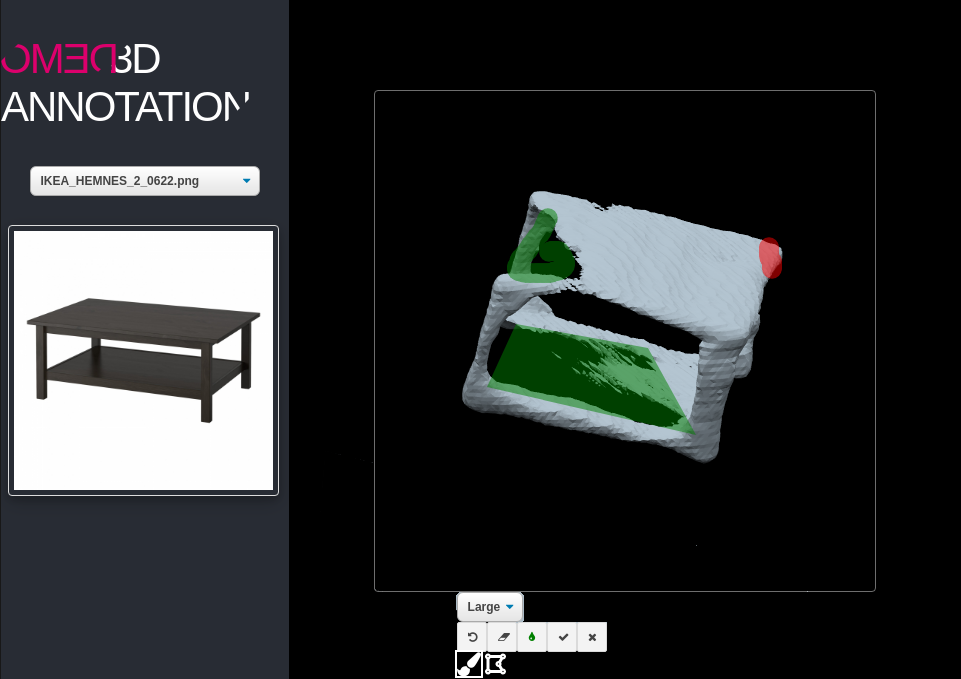}
  \caption{Screenshot.}
  \label{fig:tool}
\end{wrapfigure}
To evaluate our method with real human annotators, we built a web tool visualized in Fig.~\ref{fig:tool}, and conducted a user study on two real-world datasets: Pix3D~\cite{pix3d} and ScanNet~\cite{ScanDataset16}. The annotator engaged in the user study had no prior experience in 3D modeling and did not have access to ground truth shapes. The tool only uses scribble corrections with SIM. Details can be found in appendix. 


\subsubsection{Full Scene Annotation:}
We show full scene annotation capability of our approach on ScanNet~\cite{ScanDataset16}. 
We compare with CAD alignment in Fig.~\ref{fig:scannet} (Scan2CAD provides human-annotated CAD alignment as ground truth). Annotators took 2.48 minutes on avg. to find the appropriate CAD model and align it in the scene. We report an avg. time of 1.88 minutes in Tab.~\ref{tab:user}. Our model faithfully reconstructs the surface of all objects in the scene, aligning better with the partial surface from the scan than  CAD 
retrieval. 

\begin{minipage}{\textwidth}
  \begin{minipage}[b]{0.46\textwidth}
  \resizebox{1\textwidth}{!}{
      \begin{tabular}{|l|l|l|l|l|l|l|l|}
      \hline
                     & Bed    & Bookcase & Desk   & Sofa   & Table  & Wardrobe & Avg.   \\ \hline
      Avg. \# Scrib. & 15     & 18       & 10     & 10     & 19     & 9        & 14     \\ \hline
      Avg. Time(sec) & 141    & 109      & 116    & 71     & 184    & 55       & 113    \\ \hline
      CD Before      & 0.1555 & 0.1048   & 0.1330 & 0.0893 & 0.1637 & 0.1043   & 0.1251 \\ \hline
      CD After       & 0.0859 & 0.0517   & 0.0778 & 0.0746 & 0.0989 & 0.0513   & 0.0733 \\ \hline
      \end{tabular}
  }
    \captionof{table}{ Statistics on Pix3D.}
    \label{tab:user}
  \end{minipage}
  \hfill
  \begin{minipage}[b]{0.53\textwidth}
    \resizebox{1\textwidth}{!}{
    \begin{tabular}{|l|l|l|l|l|l|l|l|}
    \hline
                  & Bed    & Bookcase & Desk   & Sofa   & Table  & Wardrobe & Avg.   \\ \hline
    Pretrained     & 0.1728 & 0.1603   & 0.2581 & 0.0998 & 0.1659 & 0.1863   & 0.1738 \\ \hline
    Tuning on Ours & 0.1094 & 0.1017   & 0.1642 & 0.0870 & 0.1279 & 0.0898   & 0.1133 \\ \hline
    Tuning on GT   & 0.0712 & 0.0776   & 0.1168 & 0.0800 & 0.1023 & 0.0555   & 0.0839 \\ \hline
    \end{tabular}%
    }
    \captionof{table}{ Reconstruction results of OccNet on Pix3D before and after fine-tuning. }
    \label{tab:fine_tuning}
  \end{minipage}
\end{minipage}

\begin{minipage}{\textwidth}
  \hfill
  \begin{minipage}[b]{0.4\textwidth}
    \centering
      \includegraphics[width=0.7\linewidth]{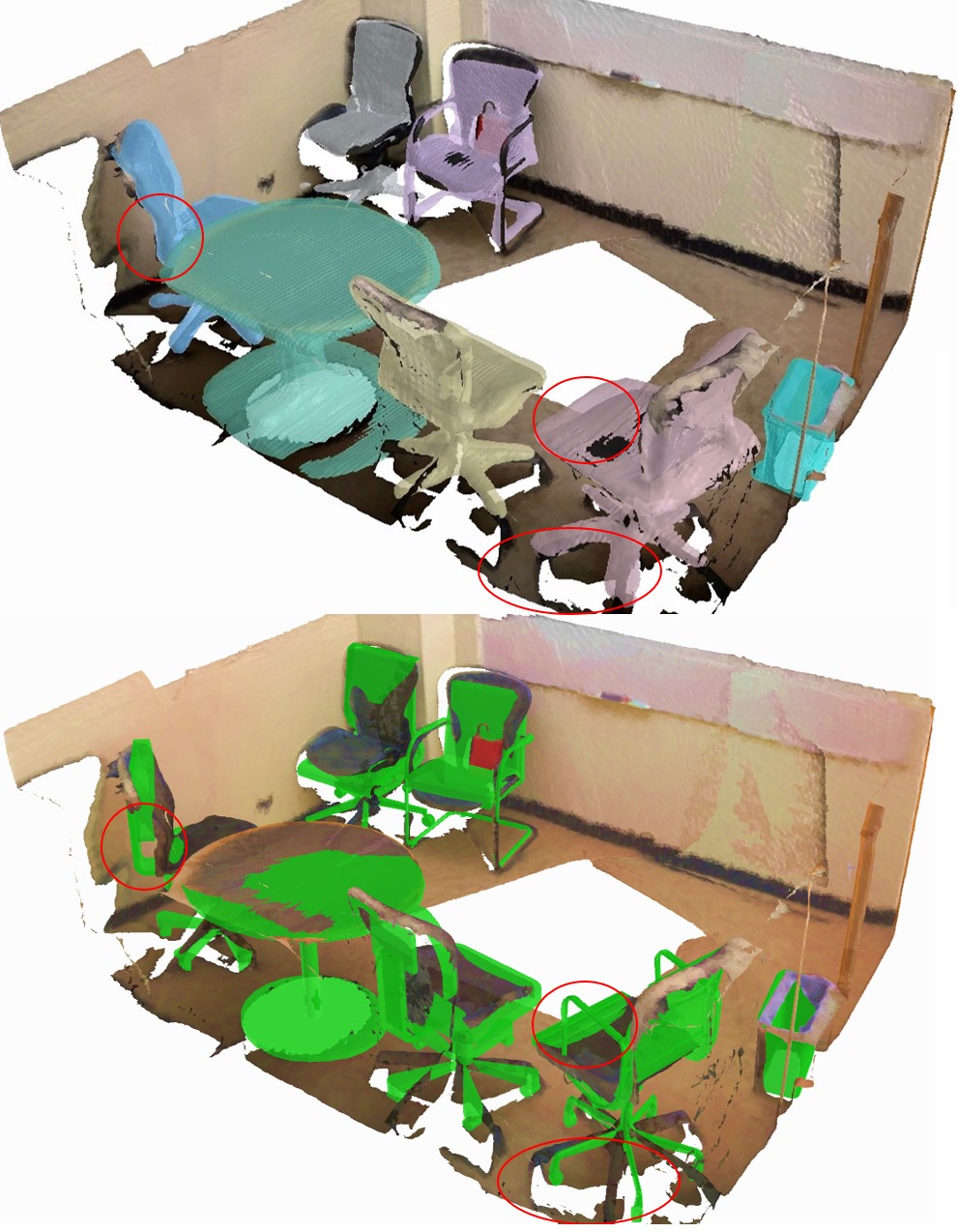}
  \captionof{figure}{ Full scene annotation (top) vs. CAD alignment (bottom) on ScanNet.}
  \label{fig:scannet}
\end{minipage}
\hfill
\begin{minipage}[b]{0.55\textwidth}
  \includegraphics[width=1\linewidth,trim=0 20 0 15,clip]{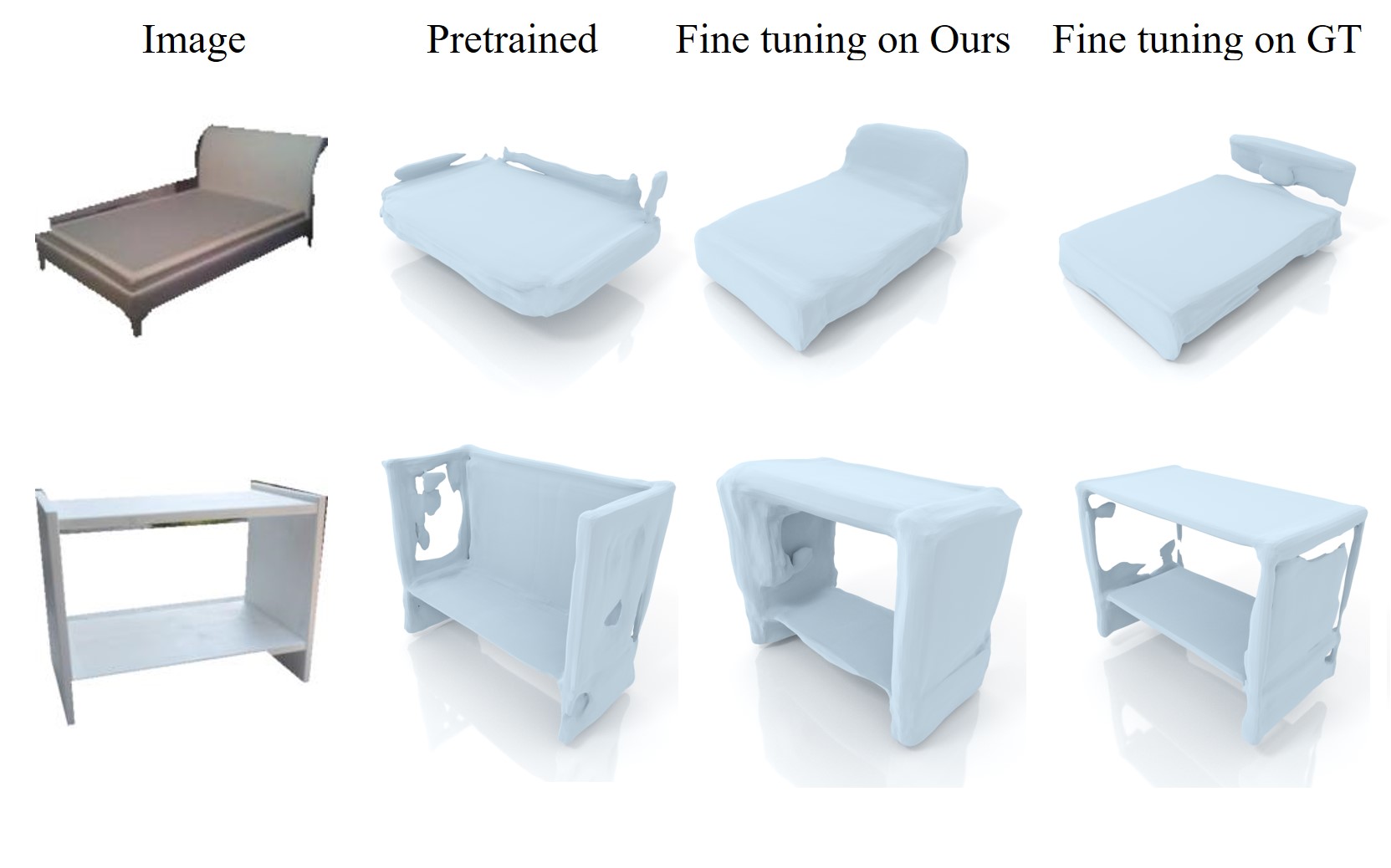}

    \captionof{figure}{ Examples showing fine-tuning OccNet with shapes annotated by our tool can effectively improve the quality of reconstruction on target task. Note that the \textit{bed} category (top) is unseen in the pretrained OccNet model.}
    \label{fig:tuning_qualitative}
  \end{minipage}
  \hfill

\end{minipage}

\vspace{-3mm}
\subsubsection{Real Dataset Creation:} To validate if our annotated data is useful for learning, we create a dataset from Pix3D images and use it to fine-tune a 3D reconstruction model pretrained on synthetic data. In this study, we chose to annotate all unseen categories in Pix3D, which has 2817 images containing un-occluded and un-truncated objects,  corresponding to 147 different CAD shapes. The annotator was asked to annotate 95 randomly-selected shapes from random views and the remaining shapes were used for testing.  
Annotation results are shown in Tab.~\ref{tab:user}. We found that the large errors in the initial prediction were effectively fixed by user corrections, but the surface is often jagged due to human error. Qualitative examples and screen recordings of the annotation process are available in the appendix. Note that runtimes of SIM and Vox2Mesh
are 0.49s and 0.007s, respectively. The majority of the time is spent on inspecting shapes and drawing scribbles. 
The improvement on Chamfer Distance is comparable to that using simulated scribbles, showing that our simulation is representative of real human inputs. In addition, we found humans can infer invisible (occluded) 3D geometry from a single image much better than the SOTA 3D reconstruction models, which is the gap that we aim to bridge with our proposed interactive annotation tool.

\begin{figure}[t]
    \begin{center}
        \includegraphics[width=0.95\linewidth]{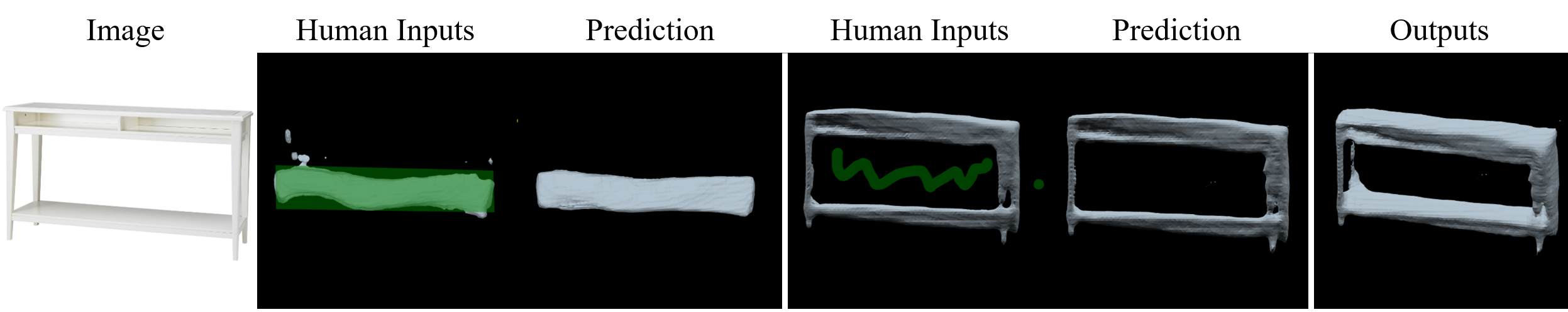}
    \end{center}
   \caption{ (left): Example of user annotating with a polygon, which traces the boundary instead of the skeleton of the negative region. (right): Example when the human provides wrong scribbles. Notice that this scribble is being largely ignored since the network tries to preserve the integrity of the shape.
}
\label{fig:human_error}

\end{figure}

We used the 95 annotated shapes with 1880 corresponding images to fine-tune OccNet~\cite{occnet}  (instead of GenRe, to avoid model bias) pre-trained on ShapeNet, and report performance in Tbl.~\ref{tab:fine_tuning} with qualitative examples in Fig.~\ref{fig:tuning_qualitative}. Despite the remaining artifacts due to annotation errors, the integrity of the reconstructed shape is largely improved, which demonstrates that our annotation tool is a step towards facilitating real-world 3D annotation for learning. Note that although SIM is trained with simulated scribbles, our model shows \textbf{robustness to human input} \ie to human annotation errors or noise (see Fig.~\ref{fig:human_error}).

\subsubsection{Limitation and Future Work:}
During our user study, we found it was difficult to annotate 1) complex shapes from bad initial predictions and 2) small but fine details. The \textit{addition} signal from the user is sometimes falsely ignored, causing repeated scribbles at the same location. Vox2Mesh sometimes creates artifacts along thin edges (\eg,chair leg, airfoil), which is commonly observed in deformation-based mesh generation methods\cite{groueix2018, pixel2mesh, meshrcnn}. We include examples of these failure cases in the appendix. To further improve our model, a possible direction is to apply the interaction in other form of representations, such as implicit function, instead of volumetric space to improve the quality of the generated shape, and avoid the additional step of converting voxels to a mesh. 



%% file: conclude.tex
\section{Conclusion}

In this paper, we provide a 3D annotation framework with two simple-to-use interaction methods, where user can correct 3D shape with scribbles and drag-and-drop clicks, and our model takes these feedback and refines 3D prediction. Experiments on both synthetic images and real sensory data demonstrate the effectiveness of our method. Our tool is available as a webservice to facilitate the progress of 3D dataset collection for 3D reasoning.


\begin{flushleft}
\textbf{Acknowledgments.} We thank Louis Clergue for assistance with developing the web tool and extended discussion. This work was supported by NSERC. SF acknowledges the Canada CIFAR AI Chair award at the Vector Institute.
\end{flushleft}